\documentclass[10pt,twocolumn,letterpaper]{article}

\usepackage{3dv}
\usepackage{times}
\usepackage{epsfig}
\usepackage{graphicx}
\usepackage{amsmath}
\usepackage{amssymb}

\usepackage{soul}
\usepackage{xcolor}

\newcommand{\revised}[1]{{\color{black} #1}}

\graphicspath{
    {./img_reduced/}
    }

\usepackage[pagebackref=true,breaklinks=true,letterpaper=true,colorlinks,bookmarks=false]{hyperref}

\colorlet{LinkCol}{red!80!black}%
\colorlet{CiteCol}{blue}%
\colorlet{CatCol}{red!40!black}
\hypersetup{
	colorlinks, linkcolor={LinkCol},
	citecolor={CiteCol}, urlcolor={blue}
}

\threedvfinalcopy 

\def\threedvPaperID{219} 

\newcommand{\mypara}[1]{\vspace{2mm}\noindent\textbf{\textit{#1}}}

\ifthreedvfinal\fi
\begin{document}

\title{Effective Rotation-invariant Point CNN with Spherical Harmonics Kernels}

\author{
Adrien Poulenard\\
LIX, Ecole Polytechnique\\
{\tt\small adrien.poulenard@inria.fr}
\and
Marie-Julie Rakotosaona\\
LIX, Ecole Polytechnique\\
{\tt\small mrakotos@lix.polytechnique.fr}
\and
Yann Ponty\\
LIX, Ecole Polytechnique\\
{\tt\small yann.ponty@lix.polytechnique.fr}
\and
Maks Ovsjanikov\\
LIX, Ecole Polytechnique\\
{\tt\small maks@lix.polytechnique.fr}
}

\maketitle

\begin{abstract}
 We present a novel rotation invariant architecture operating directly
on point cloud data. We demonstrate how rotation invariance can be
injected into a recently proposed point-based PCNN architecture, on
all layers of the network. This leads to invariance to both global shape
transformations, and to local rotations on the level of
patches or parts, useful when dealing with non-rigid objects. We
achieve this by employing a spherical harmonics-based kernel at different
layers of the network, which is guaranteed to be invariant to rigid
motions. We also introduce a more efficient pooling operation for
PCNN using space-partitioning data-structures. This results in a
flexible, simple and efficient architecture that achieves accurate
results on challenging shape analysis tasks, including classification
and segmentation, without requiring data-augmentation typically
employed by non-invariant approaches\footnote{Code and
data are provided on the project page \url{https://github.com/adrienPoulenard/SPHnet}.}.
\end{abstract}

\section{Introduction}
Analyzing and processing 3D shapes using deep learning approaches has recently attracted a lot of attention, inspired in
part by the success of such methods in computer vision and other fields.  While early approaches in this area relied on methods developed in the image domain, e.g. by sampling 2D views around the 3D object \cite{su2015multi}, or using volumetric convolutions \cite{wu20153d}, recent methods have tried to directly exploit the 3D structure of the data. This notably includes both mesh-based approaches that operate on the surface of the shapes
\cite{masci2015geodesic,monti2017geometric,poulenard2018multi}, and point-based techniques that only rely on the 3D coordinates of the shapes without requiring any connectivity information \cite{qi2017pointnet,qi2017pointnetpp}.

Point-based methods are particularly attractive, being both very general and often more efficient, as they do not require maintaining complex and expensive data-structures, compared to volumetric or mesh-based methods. As a result, starting from the seminal works of PointNet~\cite{qi2017pointnet} and PointNet++~\cite{qi2017pointnetpp}, many point-based learning approaches have been proposed, often achieving remarkable accuracy in tasks such as shape classification and segmentation among many others. A key challenge when applying these methods in practice, however, is to ensure \emph{invariance} to different kinds of transformations, and especially to rigid motions. Common strategies include either using spatial transformer blocks \cite{jaderberg2015spatial} as done in the original PointNet architecture and its extensions, or applying extensive \emph{data augmentation} during training to learn invariance from the data. Unfortunately, when applied to shape collections that are not pre-aligned, these solutions can be expensive, requiring unnecessarily long training. Moreover, they can even be incomplete when \emph{local} rotation invariance is required, e.g. for non-rigid shapes, undergoing articulated motion, which is difficult to model through data augmentation alone. 

In this paper, we propose a different approach for dealing with both global and local rotational invariance for point-based 3D shape deep learning tasks. Instead of learning invariance from data, we propose to use a different kernel that is theoretically guaranteed to be invariant to rotations, while remaining informative. 
To achieve this, we leverage the recent PCNN by extension operators \cite{atzmon2018point}, which provides an efficient framework for point-based convolutions. We extend this approach by introducing a rotationally invariant kernel and making several modifications for improved efficiency. We demonstrate on a range of difficult experiments that our method can achieve high accuracy directly, without relying on data augmentation.

\section{Related work}
A very wide variety of learning-based techniques have been proposed for 3D shape analysis and processing. Below we
review methods most closely related to ours, focusing on point-based approaches, and various ways of
incorporating rotational invariance and equivariance in learning. We refer the interested readers to several recent
surveys, e.g.  \cite{xu2016data,monti2017geometric}, for an in-depth overview of 
geometric deep learning methods.

\mypara{Learning in Point Clouds.} Learning-based approaches, and especially those based on deep learning, have recently
been proposed specifically to handle point cloud data. The seminal PointNet architecture \cite{qi2017pointnet} has
inspired a large number of extensions and follow-up works, notably including PointNet++ \cite{qi2017pointnetpp} and Dynamic Graph CNNs \cite{wang2018dynamic} for shape classification and segmentation. More recent works include PCPNet
\cite{guerrero2018pcpnet} for normal and curvature estimation, PU-Net \cite{yu2018pu} for point cloud
upsampling, and PCN for shape completion \cite{yuan2018pcn} among many others.

While the original PointNet approach is not based on a convolutional architecture, instead using a series of classic
MLP fully connected layers, several methods have also tried to define and exploit meaningful notions of convolution on point
cloud data, inspired by their effectiveness in computer vision. Such approaches notably include: basic pointwise convolution through nearest-neighbor binning and a grid kernel \cite{hua2018pointwise}; Monte Carlo convolution, aimed at dealing
with non-uniformly sampled point sets \cite{hermosilla2018monte}; learning an $\mathcal{X}$-transformation of the input
point cloud, which allows the application of standard convolution on the transformed representation \cite{li2018pointcnn}; and using extension operators for applying point convolution \cite{atzmon2018point}. These techniques primarily
differ by the notion of neighborhood and the construction of the kernels used to define convolution on the point clouds. Most of them, however,
share with the PointNet architecture a lack of support for invariance to rigid motions, mainly because their kernels are applied to point coordinates, and defining invariance at the level of individual points is generally not meaningful.

\mypara{Invariance to transformations.} Addressing invariance to various transformation classes has been considered in
many areas of Machine Learning, and Geometric Deep Learning in particular. Most closely related to ours are approaches
based on designing \emph{steerable filters}, which can learn representations that are equivariant to the rotation of the
input data \cite{weiler2018learning,weiler20183d,andrearczyk2018exploring,worrall2018cubenet}. A particularly comprehensive overview of the key ideas and results in this area is presented in
\cite{kondor2018generalization}. In closely related works, Cohen and colleagues have proposed group equivariant networks \cite{cohen2016group} and rotation-equivariant spherical CNNs \cite{cohen2018spherical}. While the theoretical foundations of these approaches are well-studied in the context of 3D shapes, they have primarily been applied to either volumetric \cite{weiler20183d} or spherical (e.g. by projecting shapes onto an enclosing sphere via ray casting) \cite{cohen2018spherical} representations. Instead, we apply these constructions directly in an efficient and powerful \emph{point}-based architecture. 

Perhaps most closely related to ours are two very recent unpublished methods, \cite{you2018prin,thomas2018tensor}, that also aim to introduce invariance into point-based networks. Our approach is different from both, since unlike the PRIN method in \cite{you2018prin} our convolution operates directly on the point clouds, thus avoiding the construction of spherical voxel space. As we show below, this gives our method greater invariance to rotations and higher accuracy. At the same time while the authors of \cite{thomas2018tensor} explore similar general ideas and describe related constructions, including the use of spherical harmonics kernels, they do not describe a detailed  architecture, and only show results with dozens of points (the released implementation is also limited to toy examples), rendering both the method and its exact practical utility unclear. Nevertheless,
we stress that both \cite{you2018prin} and \cite{thomas2018tensor} are very recent unpublished methods and thus concurrent to our approach.


\mypara{Contribution:} Our key contributions are as follows:
\begin{enumerate}
    \setlength\itemsep{0.25em}
\item We develop an effective rotation invariant point-based network. To the best of our knowledge, ours is the first such method achieving higher accuracy than PointNet++ \cite{qi2017pointnetpp} \emph{with data augmentation} on a range of tasks.
\item We significantly improve the efficiency of PCNN by Extension Operators \cite{atzmon2018point} using space partitioning. 
\item We demonstrate the efficiency and accuracy of our method on tasks such as shape classification, segmentation and matching on standard benchmarks and introduce a novel dataset for RNA molecule segmentation.
\end{enumerate}

\section{Background}

In this section we first introduce the main notation and give a brief overview of the PCNN approach \cite{atzmon2018point} that we use as a basis for our architecture.  

\subsection{Notation}

We use the notation from \cite{atzmon2018point}.  In particular, we use tensor notation: $a \in \mathbb{R}^{I \times I \times J \times L \times M}$ and the sum of tensors $c = \sum_{ijl} a_{ii'jlm}b_{ijl}$ is defined by the free indices: $c = c_{i'm}$.  $ C(\mathbb{R}^3, \mathbb{R}^K)$ represent the collections of volumetric functions $\psi : \mathbb{R}^3 \rightarrow \mathbb{R}^K$.

\subsection{The PCNN framework}
The PCNN framework consists of three simple steps. First a signal is extended from a point cloud to  $\mathbb{R}^3$  using an extension operator $\mathcal{E}_X$. Then standard convolution on volumetric functions  $O$ is applied. Finally, the output is restricted to the original point cloud with a restriction operator  $\mathcal{R}_X$. The final convolution on point clouds is defined as:
\begin{equation} \label{eq:pc_conv_op}
    O_X = \mathcal{R}_X \circ O \circ \mathcal{E}_X
\end{equation}

In  the original work \cite{atzmon2018point}, the extension operators and kernel functions are chosen so that the composition of the three operations above, using  Euclidean convolution in $\mathbb{R}^3$ can be computed in closed form.

\paragraph{Extension operator.} Given an input signal represented as $J$ real-valued functions $f \in \mathbb{R}^{I\times J}$ defined on a point cloud, it can be extended to $\mathbb{R}^3$ via a set of volumetric basis functions $l_i \in C(\mathbb{R}^3, \mathbb{R})$ using the values of $f$ at each point $f_i$. 
The extension operator $ \mathcal{E}_X : \mathbb{R}^{I \times J} \rightarrow C(\mathbb{R}^3, \mathbb{R}^J)$ is then: 
\begin{equation}
    (\mathcal{E}_X[f])_j(x) =  \sum_i f_{ij}l_i(x)
\end{equation}
The authors of \cite{atzmon2018point} use Gaussian basis functions centered at the points of the point cloud so that the number of basis functions equals the number of points.
\paragraph{Convolution operator.} Given a kernel $\kappa \in C(\mathbb{R}^3, \mathbb{R}^{J \times M}) $, the convolution operator $O : C(\mathbb{R}^3, \mathbb{R}^J) \rightarrow C(\mathbb{R}^3, \mathbb{R}^M)$ applied to a volumetric signal $\psi \in C(\mathbb{R}^3, \mathbb{R}^J)$ is defined as: 
\begin{equation}
    (O[\psi])_m(x) = (\psi * \kappa)_m(x)=  \int_{\mathbb{R}^3} \sum_j \psi_j(y) \kappa_{jm}(x- y) dy
\end{equation}
The kernel can be represented in an RBF basis: 
\begin{equation}
    \kappa_{jm}(z) = \sum_l k_{jml} \Phi(|z - v_l|), 
\end{equation}
where $k_{jml}$ are learnable parameters of the network, $\Phi$ is the Gaussian kernel and ${v}_{l=1}^L$ represent translation vectors in $\mathbb{R}^3$. For instance, they can be chosen to cover a standard $3 \times 3 \times 3$ grid.  
\paragraph{Restriction operator.} The restriction operator $R_X : C(\mathbb{R}^3, \mathbb{R}^J) \rightarrow \mathbb{R}^{I\times J}$ is defined as simply as the restriction of the volumetric signal to the point cloud:
\begin{equation} \label{eq:PCNN_restriction}
    (R_X[\psi])_{i,j} = \psi_j(x_i)
\end{equation}

\paragraph{Architecture}
With these definitions in hand, the authors of \cite{atzmon2018point} propose to stack a series of convolutions with non-linear activation and pooling steps into a robust and flexible deep neural network architecture showing high accuracy on a range of tasks.

\section{Our approach}

\begin{figure}[t!]
    \centering
     \includegraphics[width=\columnwidth]{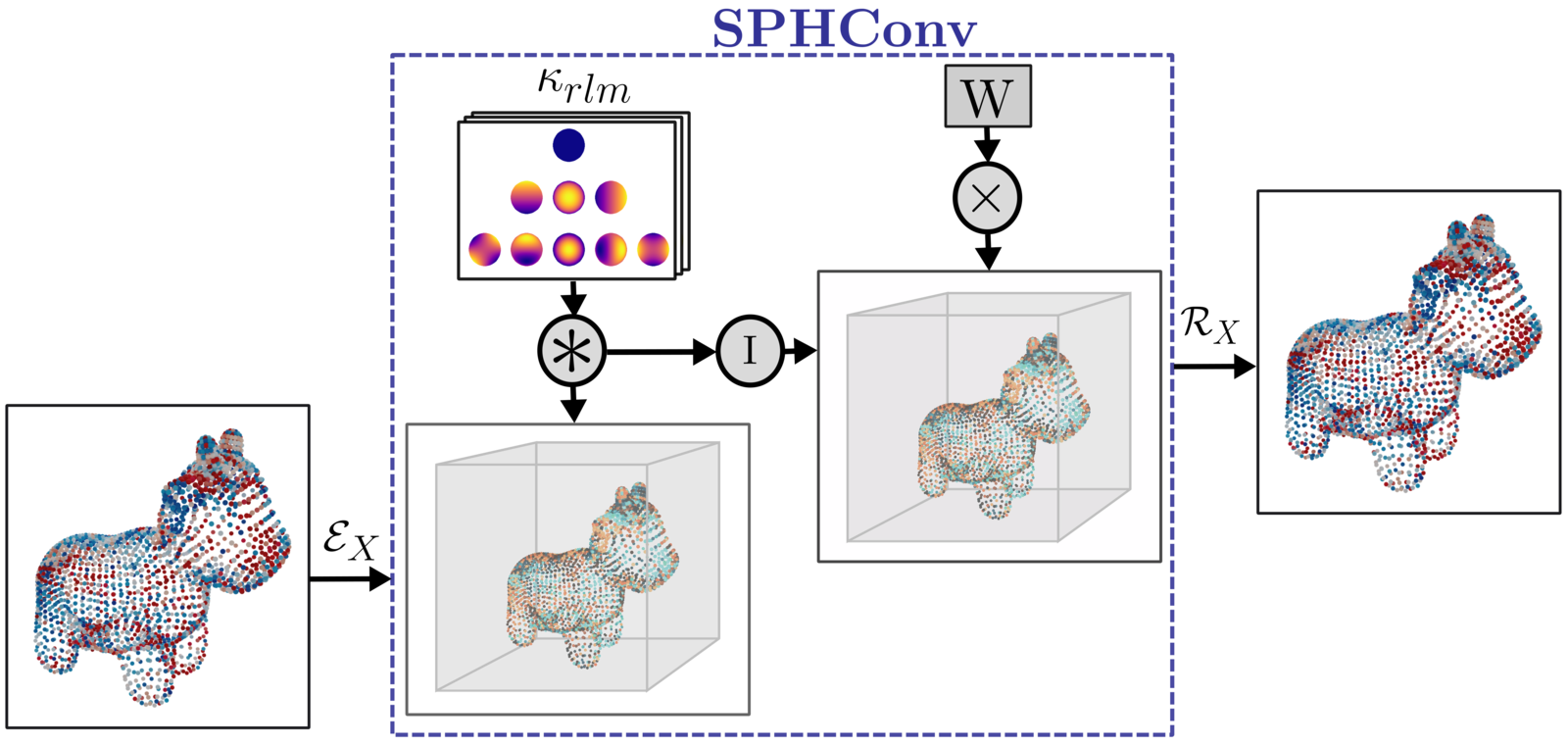}
    \caption{SPHNet framework. A signal on point cloud is first extended to $\mathbb{R}^3$ (left). Our convolution operator is applied to the extended function (center). The signal is restrained to the original point cloud (right). 
    \label{fig:overview}\vspace{-2mm}}
\end{figure}

\paragraph*{Overview}
 Our main goal is to extend the PCNN approach to develop a rotation invariant convolutional neural network on point clouds. We call our network SPHNet due to the key role that the spherical harmonics kernels play in it. Figure \ref{fig:overview} gives an overview. Following PCNN we first extend a function on point clouds to a volumetric function by the operator $\mathcal{E}_X$. Secondly, we apply the convolution operator SPHConv to the volumetric signal. Finally, the signal is restricted by $R_X$ to the original point cloud.


\subsection{Spherical harmonics kernel} \label{spherical_harmonics_kernel}
In this work, we propose to use spherical harmonics-based kernels to design a point-based rotation-invariant network. In \cite{weiler20183d}, the authors define spherical harmonics kernels with emphasis on rotation equivariance, applied to volumetric data. We adopt these constructions to our setting to define rotation-invariant convolutions.

Spherical harmonics is a family of real-valued functions on the unit sphere which can be defined, in particular, as the eigenfunctions of the spherical Laplacian. Namely, the $\ell^{th}$  spherical harmonic space has dimension $2\ell+1$ and is spanned by spherical harmonics $(Y_{\ell,m})_{\ell>=0,m \in {-\ell...\ell} }$, where $\ell$ is the degree. Thus, each $Y_{\ell,m}$ is a real-valued function on the sphere $Y_{\ell,m}: \mathcal{S}_2 \rightarrow \mathbb{R}.$

Spherical harmonics are rotation equivariant. For any rotation $R \in \mathrm{SO}(3)$ we have:
\begin{equation}
Y_{\ell, m}(R x) = \sum_{n = -\ell}^{\ell} D^{\ell}_{mn}(R) Y_{\ell, n}(x).
\end{equation}
Where $D^{\ell}(R)$ is the so-called Wigner matrix of size $2\ell+1 \times 2\ell+1$, \cite{wigner1959group}.   \revised{
  Importantly, Wigner matrices are \emph{orthonormal} for all $\ell$ making the norm of every spherical harmonic space invariant to rotation. This classical fact has been exploited, for example in \cite{kazhdan2003rotation} to define global rotation-invariant shape descriptors. More generally the idea of using bases of functions having a predictable behaviour under rotation to define invariant or equivariant filters or descriptors is closely related to the classical concept of steerable filters \cite{freeman1991design}. }


The spherical harmonic kernel basis introduced in \cite{weiler20183d} is defined as: 
\begin{align}
\kappa_{r\ell m}(x) = \exp\left(-\frac{\left|\|x\|_2 - \rho \frac{r}{n_R-1} \right|^2}{2\sigma^2} \right) Y_{\ell, m}\left(\frac{x}{\|x\|_2}\right),
\end{align}
where, $\rho$ is a positive scale parameter, $n_{R}$ is the number of radial samples and $\sigma = \frac{\rho}{n_R-1}$. Note that the kernel depends on a radial component, indexed by $r \in 0...n_R-1$, and defined by Gaussian shells of radius $r \frac{\rho}{n_R-1}$, and an angular component indexed by $\ell, m$ with $\ell \in 0...n_{L}-1$ and $m \in -\ell...\ell$, defined by values of the spherical harmonics. 

 This kernel inherits the behaviour of spherical harmonics under rotation, that is:
\begin{equation} \label{eq:rot_D}
\kappa_{r\ell m}(R x ) = \sum_{n = -\ell}^{\ell} D^{\ell}_{mn}(R)\kappa_{r \ell n}(x)
\end{equation}
where $R \in SO(3)$. 

\subsection{Convolution layer} \label{convolution_layer}


Below, we describe our SPHNet method. To define it we need to adapt the convolution and extension operators used in PCNN \cite{atzmon2018point}, while the restriction operators are kept exactly the same.

\paragraph{Extension.} PCNN uses Gaussian basis functions to extend functions to $\mathbb{R}^3$. However,  convolution of the spherical harmonics kernel with Gaussians does not admit a closed-form expression. Therefore, we ``extend'' a signal $f$ defined on a point cloud via a weighted combination of Dirac measures. Namely $\mathcal{E}_X : \mathbb{R}^{I \times J} \rightarrow C(\mathbb{R}^3, \mathbb{R}^J)$ is:
\begin{equation} 
(\mathcal{E}_X[f])_j = \sum_i f_{ij} \omega_i \delta_{x_i},
\end{equation}
where we use the weights: $\omega_i = 1 / (\sum_j \exp(-\frac{||x_j - x_i ||^2}{2\sigma^2}))$. The Dirac measure has the following property:
\begin{equation} \label{eq:dirac_int}
    \int _{X}f(y)\delta _{x}(y)\,\mathrm {d} y=f(x)
\end{equation}

 \paragraph{Convolution. } We first introduce a non-linear rotation-invariant convolution operator: SPHConv. 
Using Eq. \eqref{eq:dirac_int}, the convolution between an extended signal and the spherical harmonic kernel $S[f] : C(\mathbb{R}^3, \mathbb{R}^J) \rightarrow C(\mathbb{R}^3, \mathbb{R}^{J})$ is given by:
\begin{equation}
\label{eq:def_convolution}
    (S[f])_j(x) = (\mathcal{E}_X[f] \ast \kappa_{r\ell m})_j(x) 
= \sum_i f_{ij} \omega_i \kappa_{r\ell m}(x_i - x)
\end{equation}

 Using Eq. \eqref{eq:rot_D}, we can express the convolution operator when a rotation $R$ is applied to the point cloud $X$ as a function of the kernel functions: 
\begin{align}
\nonumber
  &(R(\mathcal{E}_X[f] \ast \kappa_{r\ell m}))_j(x) = 
\sum_i f_{ij} \omega_i \kappa_{r\ell m}(R(x_i - x)) \\ 
\nonumber
& =  \sum_{n=-\ell}^{\ell}  D^{\ell}_{mn}(R) \sum_i f_{ij} \omega_i \kappa_{r\ell n}(x_i - x)\\ 
& = \sum_{n=-\ell}^{\ell} D^{\ell}_{mn}(R)(\mathcal{E}_X[f] \ast \kappa_{r\ell n})_j(x)
\label{eq:conv_props}
\end{align}

 We observe that a rotation of the point cloud induces a rotation in feature space. In order to ensure rotation invariance we recall that the Wigner matrices $D^l$ are all orthonormal. Thus, by taking the norm of the convolution with respect to the degree of the spherical harmonic kernels, we can gain independence from $R$. To see this, note that thanks to Eq. \eqref{eq:conv_props} only the $m$-indexed dimension of $(\mathcal{E}_X[f] \ast \kappa_{r\ell m})_j(x)$ is affected by the rotation. Therefore, we define our rotation-invariant convolution operator
$I_{rl}[X,f] : C(\mathbb{R}^3, \mathbb{R}^J) \rightarrow C(\mathbb{R}^3, \mathbb{R}^{J}) $ as:
 \begin{equation}
   (I_{rl}[X, f])_j(x) =  \|(\mathcal{E}_X[f] \ast \kappa_{r\ell m})_j(x) \|_2^m,
\end{equation}
where for a tensor $T_{rlmj}$ we use the notation $\|T\|_2^m$ to denote a tensor $T^*$ obtained by taking the $L_2$ norm along the $m$ dimension: $T^*_{rlj} = \sqrt{\sum_m T_{rlmj}^2} $.

Importantly, unlike the original PCNN approach \cite{atzmon2018point}, we cannot apply learnable weights directly on $(\mathcal{E}_X[f] \ast \kappa_{r\ell m})_j(x)$ as the result would not be rotation invariant. Instead, we  take a linear combination of $(I_{rl}[X, f])$, obtained after taking the reduction along the $m$ dimension above, using learnable weights $W \in \mathbb{R}^{G \times I \times n_R \times r_L}$, where $G$ is the number of output channels. This leads to:
\begin{equation} \label{eq:mult_weights}
(O[f])_g(x) = \xi \left(\sum_{j r \ell } W_{gjr \ell }(I_{rl}[X, f])_j(x) + b_g \right)
\end{equation}

Finally, we define the convolution operator $O_X: C(\mathbb{R}^3, \mathbb{R}^J) \rightarrow C(\mathbb{R}^3, \mathbb{R}^{G}) $ by restricting the result  to the point cloud as in Eq. \eqref{eq:PCNN_restriction}:
\begin{equation}
    ((O_X)_g[f])_i = \xi \left(\sum_{j r \ell} W_{gjr \ell }(I_{rl}[X, f](x_i) + b_g \right)
\end{equation}


\subsection{Pooling and upsampling}


In addition to introducing a rotation-invariant convolution into the PCNN framework, we also propose several improvements, primarily for computational efficiency. 
 
 The original PCNN work \cite{atzmon2018point} used Euclidean farthest point sampling and pooling in Voronoi cells. Both of these steps can be computationally inefficient, especially when handling large datasets and with data augmentation. Instead, we propose to use a space-partitioning data-structure for both steps using constructions similar to those in \cite{klokov2017escape}. For this, we start by building a kd-tree for each point cloud, which we then use as follows.
 

 \vspace{-1mm}
 \paragraph{Pooling. } For a given point cloud $P$, our pooling layer of depth $k$ reduces $P$ from size $2^n$ to $2^{n-k}$ by applying max pooling to the features of each subtree. The coordinates of the points of the subtree are averaged. The resulting reduced point cloud kd-tree structure and indexing can be immediately computed from the one computed for $P$. This gives us a family ${T}_{k \in 1...n}$ of kd-trees of varying depths.    

 \vspace{-1mm}
 \paragraph{Upsampling. } The upsampling layer is computed simply by repeating the features of each point of a point cloud at layer $k$ using the kd-tree of structure $T_k$. The upsampled point cloud follows the structure of $T_{k+1}$.

  \vspace{-1mm}
\paragraph{Comparison to PCNN.} In PCNN pooling is performed through farthest point sampling. The maximum over the corresponding Voronoi cell is then assigned to each point of the sampled set. This method has a complexity of $O(|P|^2)$ while ours has a complexity of $O(|P| \log^2 |P| )$, leading to noticeable improvement in practice.  

We remark that kd-tree based pooling breaks full rotation invariance of our architecture, due to the construction of axis-aligned separating hyperplanes. However, as we show in the results below, this has a very mild effect in practice and our approach has very similar performance regardless of the rotation of the data-set. \revised{A possible way to circumvent this issue would be to modify the kd-tree construction by splitting the space along local PCA directions.}



\section{Architecture and implementation details}

We adapted the classification and segmentation architectures from \cite{atzmon2018point}. Using these as generic models we derive three general networks: our main rotation-invariant architecture $\mathbf{SPHnet}$, which stands for Spherical Harmonic Net and uses the rotation invariant convolution we described. We also compare it to two baselines: $\mathbf{SPHBase}$ is identical to $\mathbf{SPHnet}$, except that we do not take the norm of each spherical harmonic component and apply the weights directly instead. We use this as the closest non-invariant baseline. We also compare to PCNN (mod), which is also non rotation-invariant. It uses the Gaussian kernels from the original PCNN, but employs the same architecture and pooling as we use in  $\mathbf{SPHnet}$ and $\mathbf{SPHBase}$.

The original architectures in \cite{atzmon2018point} consist of arrangements of convolution blocks, pooling and up-sampling layers that we replaced by our own. We kept the basic structure described bellow. In all cases, we construct the convolutional block using one convolutional layer followed by a batch normalization layer and a ReLU non linearity. Our convolution layer depends on the following parameters:
\begin{itemize}
  \setlength\itemsep{0em}
    \item Number of input and output channels
    \item Number $n_L$ of spherical harmonics spaces in our kernel basis
    \item Number $n_R$ of spherical shells in our kernel basis
    \item The kernel scale $\rho > 0$
\end{itemize}
For the sake of efficiency we also restrict the computation of convolution to a fixed number of points using $k$-nearest neighbor patches. \revised{Note that unlike other works, e.g. \cite{wang2018dynamic} we do not learn an embedding for the intermediate features, as they are always defined on the original point cloud. Thus, to ensure full invariance we apply our rotation invariant convolution at every layer.}

The number of input channels is deduced from the number of channels of the preceding layer. We used 64 points per patch in the classification case and 48 in the segmentation case. We fixed $n_L = 4$ and $n_R = 2$ throughout all of our experiments. The scale factor $\rho$ can be defined only for the first layer and deduced for the other ones as will be explained bellow.

\subsection{Classification}
 
\begin{figure}[h!]
    \centering
     \includegraphics[width=\columnwidth]{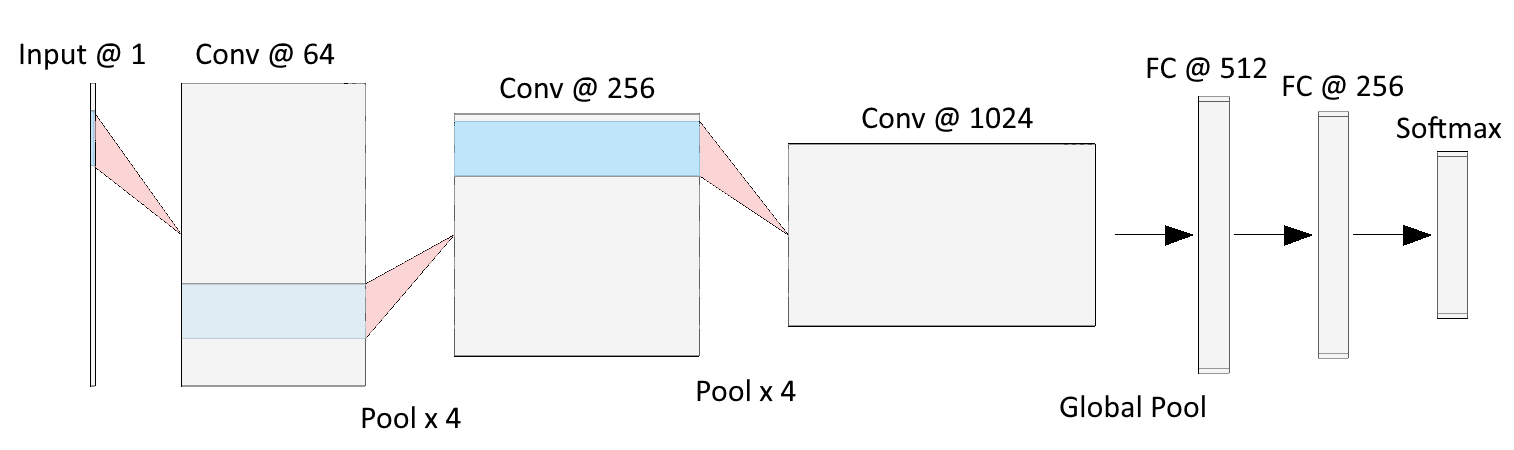}
    \caption{Our classification architecture. Conv @ $k$ indicates a conv layer with $k$ output channels, Pool $\times k$ is a pooling of factor $k$ and FC stands for fully connected layer.}
    \label{fig:classif_arch}
\end{figure}

The classification architecture is made of 3 convolutional blocks with 64, 256, 1024 output channels respectively: the first two are followed by a max pooling layer of ratio $4$, and the last one if followed by a global max pooling layer. Finally we have a fully connected block over channels composed of two layers with 512 and 256 units, each followed by a dropout layer of rate 0.5 and a final softmax layer for the classification as shown in Figure \ref{fig:classif_arch}.

We use $\rho = 0.1$ for the first layer and $2\rho$, $4\rho$ for the second and third layers. \revised{We illustrate the importance of the scale parameter $\rho$ by showing its impact on classification accuracy (see Appendix \ref{modelnet40_scale_table}).} The classification architecture we use expects a point cloud of 1024 points as input and defines the convolution layers on it according to our method. We use the constant function equal to 1 as the input feature to the first layer. Note that, since our goal is to ensure rotation invariance, we cannot use coordinate functions as input. 

\subsection{Segmentation}
\begin{figure}[h!]
\centering
     \includegraphics[width=\columnwidth]{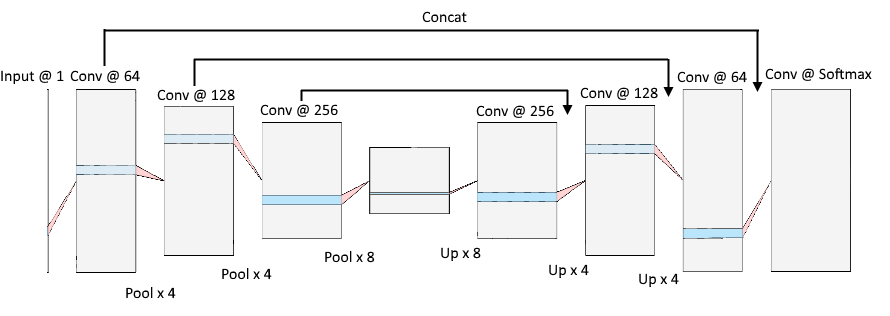}
    \caption{Segmentation architecture. Conv @ $k$ indicates a conv layer with $k$ output channels, Pool $\times k$ is a pooling of factor $k$ and Up $\times k$ is an upsampling by a factor $k$.     \label{fig:seg_arch}\vspace{-2mm} }
\end{figure}

Our segmentation network takes as input a point cloud with 2048 points and, similarly to the classification case,  we use the constant function equal to 1 as the input feature to the first layer. Our segmentation architecture is shown in Figure \ref{fig:seg_arch}.

The architecture is U-shaped consisting of an encoding and a decoding block, each having 3 convolutional blocks. The encoding convolutional blocks have 64, 128, 256 output channels respectively. The first two are followed by max pooling layers of ratio $4$ and the third by a pooling layer of ratio $8$. These are followed by a decoding block where each conv block is preceded by an up-sampling layer to match the corresponding encoding block, which is then concatenated with it. The final conv-layer with softmax activation is then applied to predict pointwise labels. The two last conv-blocks of the decode part are followed by dropout of rate 0.5. We chose a scale factor of $\rho = 0.08$ for the first convolutional layer, the two next layers in the encoding block having respective scale factors $2\rho$ and $4\rho$. The scale factors for the decode block are the same in reverse order. The scale factor for the final conv layer is $\rho$.

\section{Results}

\subsection{Classification}

We tested our method on the standard ModelNet40 benchmark \cite{wu20153d}. This dataset consists of  9843  training   and  2468  test  shapes  in  40  different classes, such as guitar, cone, laptop etc.
We use the same version as in \cite{atzmon2018point} 
with point clouds consisting of 2048 points centered and normalized so that the maximal distance of any point to the origin is 1.
We randomly sub-sample the point clouds to 1024 points before sending
them to the network. The dataset is aligned, so that all point clouds are in canonical position.

We compare the classification accuracy of our approach SPHNet with different methods for learning on point clouds in Table \ref{modelnet40_table}. In addition to the original PCNN architecture and our modified versions of it, we compare to PointNet++ \cite{qi2017pointnetpp}. We also include the results of the recent rotation-invariant framework PRIN \cite{you2018prin}.

We train all different models in two settings: we first train with the original (denoted by `O') dataset and also with the dataset augmented by random rotations (denoted by `A'). We then test with either the original testset or the testset augmented by rotations, again denoted by `O' and `A' respectively.

We observe that while other methods have a significant drop in
accuracy when trained with rotation augmentation, our method
remains stable, and in particular outperforms all methods trained and
tested with data augmentation (A/A column), implying that the orientation
of different shapes is not an important factor in the learning
process. Moreover, our method achieves the best accuracy in this
augmented training set setting.
For PRIN evaluation, we used the architecture for classification
described in \cite{you2018prin} trained for 40 epochs as suggested in
the paper. In all our experiments we train the PRIN model with the default parameters except for the bandwidth, which we set to 16 in order to fit within the 24GB of memory available in Titan RTX. As demonstrated in \cite{you2018prin} this parameter choice produces slightly lower results but they are still comparable. In our experiments, we observed that PRIN achieves poor performance when trained with rotation augmented data.

\begin{table}
\begin{center}
\begin{tabular}{|c|c|c|c|c|c|}
  \hline
  Method &  O / O &  A / O & O / A & A / A & time \\
  \hline
  PCNN   & $\mathbf{92.3}$ & 85.9 & 11.9 & 85.1 & 264 s
  \\
  PointNet++  & 91.4 & 84.0 & 10.7 & 83.4 & 78 s 
  \\
  PCNN (mod) & 91.1 & 83.4 & 9.4 & 84.5 & 22.6 s
  \\
  SPHBaseNet & 90.7 & 82.8 & 10.1 &84.8 & 25.5 s
  \\
  SPHNet (ours) & 87.7 & $\mathbf{87.1}$ & $\mathbf{86.6}$ &$\mathbf{87.6}$ & 25.5 s
  \\
  PRIN & 71.5 & 0.78 & 43.1 & 0.78 & 811 s\\
  
  \hline
\end{tabular}
\caption{Classification accuracy on the modelnet40 dataset. A stands
  for data augmented by random rotations and O for original
  data. E.g., the model A/O was trained with augmented and tested on
  the original data. Timings per epoch are given when training on a
  NVIDIA RTX 2080 Ti card. \label{modelnet40_table} \vspace{-3mm}}
\end{center}
\end{table}

\begin{figure}[t!]
    \centering
     \includegraphics[width=\columnwidth, height=145pt]{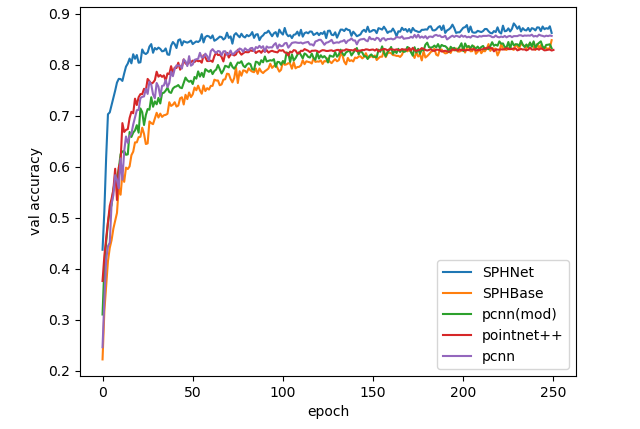}
    \caption{Validation accuracy for ModelNet40 classification with rotation augmentation at training.}
    \label{fig:modelnet40_val_acc}
\end{figure}

We also remark that even when training with data augmentation, our
method converges significantly faster and to higher accuracies since it does not need to learn invariance to rotations.  We show the evolution of the validation accuracy curves for different methods in Figure \ref{fig:modelnet40_val_acc}. 

\subsection{Segmentation}

We also applied our approach to tackle the challenging task of
segmenting molecular surfaces into  functionally-homologous regions
within RNA molecules. We considered a family of 640 structures of 5s
ribosomal RNAs (5s rRNAs), downloaded from the PDB database
\cite{berman2000protein}. A surface, or molecular envelope, was
computed for each RNA model by sweeping the volume around the atomic
positions with a small ball of fixed radius. This task was performed
using the scripting interface of the Chimera structure-modelling
environment~\cite{Pettersen2004}. Individual RNA sequences were then
globally aligned, using an implementation of the Needleman-Wunsch
algorithm~\cite{Needleman1970}, onto the RFAM~\cite{Kalvari2018}
reference alignment RF00001 (5s rRNAs). Since columns in multiple
sequence alignments are meant to capture functional homology, we
treated  each column index as a label, which we assigned to individual
nucleotides, and their corresponding atoms, within each RNA. Labels
were finally projected onto individual vertices of the surface by
assigning to a vertex the label of its closest atom. This results in each shape represented as a triangle mesh consisting of approximately 10k vertices, and its segmentation into approximately 120 regions, typically represented as connected components. 

\begin{figure}[h!]
    \centering
     \includegraphics[width=\columnwidth]{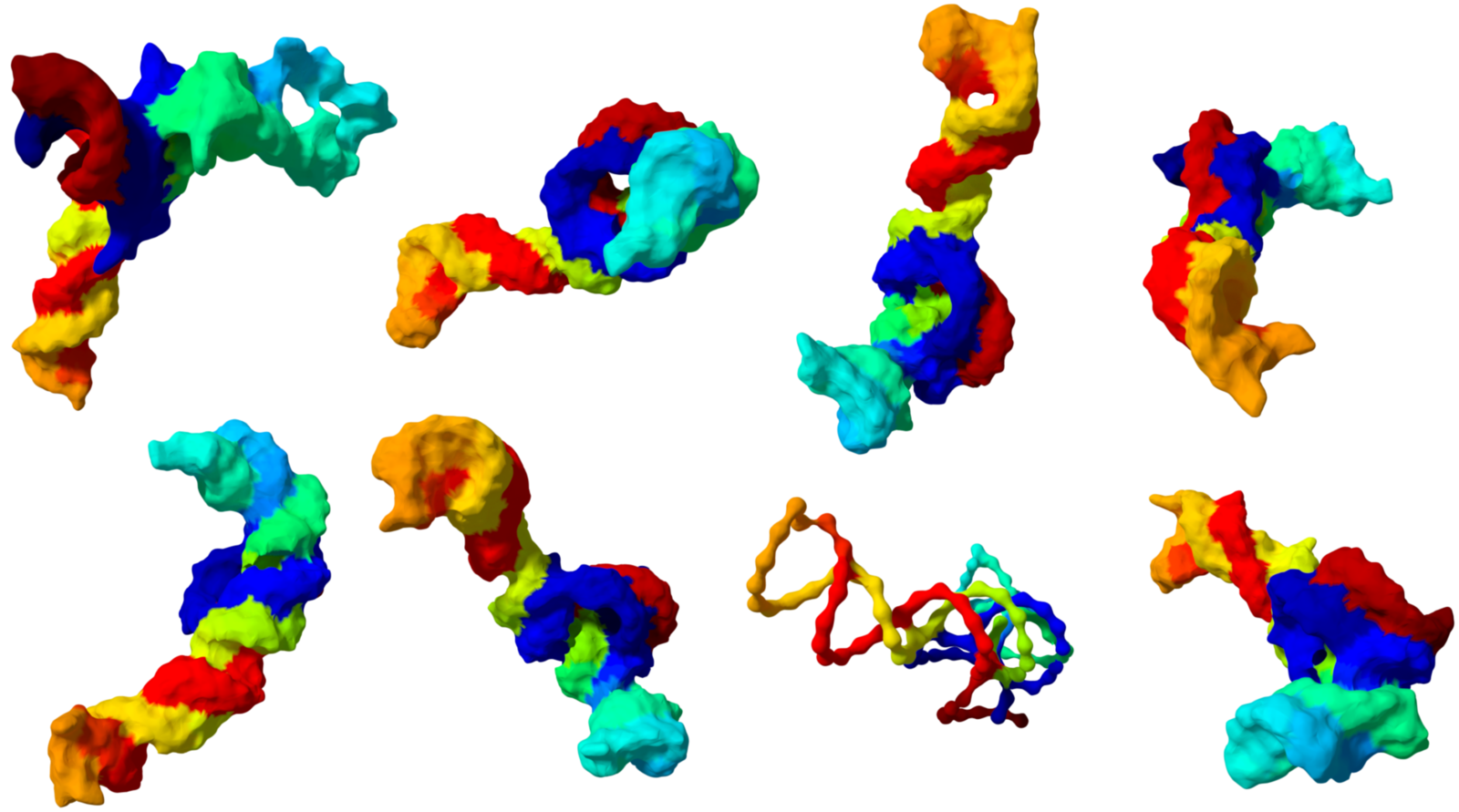}
    \caption{Labeled RNA molecules.}
    \label{fig:rna}
\end{figure}

Shapes in this dataset are not pre-aligned and can have fairly significant geometric variability arising both from different conformations as well as from the molecule acquisition and reconstruction process (see Fig. \ref{fig:rna} for an example).

Given a surface mesh, we first downsample it to 4096 points with farthest point sampling and then randomly sample to 2048 points before sending the data to the network.





We report the segmentation accuracy in Table \ref{rna_table}. As in the classification case we
observed that PRIN \cite{you2018prin} degrades severely when augmenting the training set by random
rotations. Overall, we observe that our method achieves the best accuracy in all
settings. Furthermore, similarly to the classification task, the accuracy of our method is stable in
different settings (with and without data augmentation) for this dataset.  

\begin{table}
\begin{center}
\begin{tabular}{|c|c|c|c|c|c|}
  \hline
  Method &  O/O & A/O &  O/A & A/A & time\\
  \hline
  PCNN  & 76.7 & 78.0 & 35.1 & 77.8 & 65 s 
  \\
  PointNet++  & 72.3 & 74.4 & 46.1 & 74.2 & 18 s 
  \\
  PCNN (mod) & 74.2 & 74.3 & 30.9 & 73.7 & 9.7 s
  \\
  SPHBaseNet & 74.8 & 74.7 & 28.3 & 74.8 & 17 s
  \\
  SPHNet (ours) & $\mathbf{80.8}$ & $\mathbf{80.1}$ & $\mathbf{79.5}$ &$\mathbf{80.4}$& 18 s
  \\
  PRIN & 66.9 &  6.84 &53.7 & 6.57& 10 s\\
  \hline
\end{tabular}
\caption{Segmentation accuracy on the RNA molecules dataset. Timings
  per epoch are given for an NVIDIA RTX 2080 Ti card.
\label{rna_table} \vspace{-3mm}}
\end{center}
\end{table}

\begin{figure}[t!]
    \centering
     \includegraphics[width=\columnwidth]{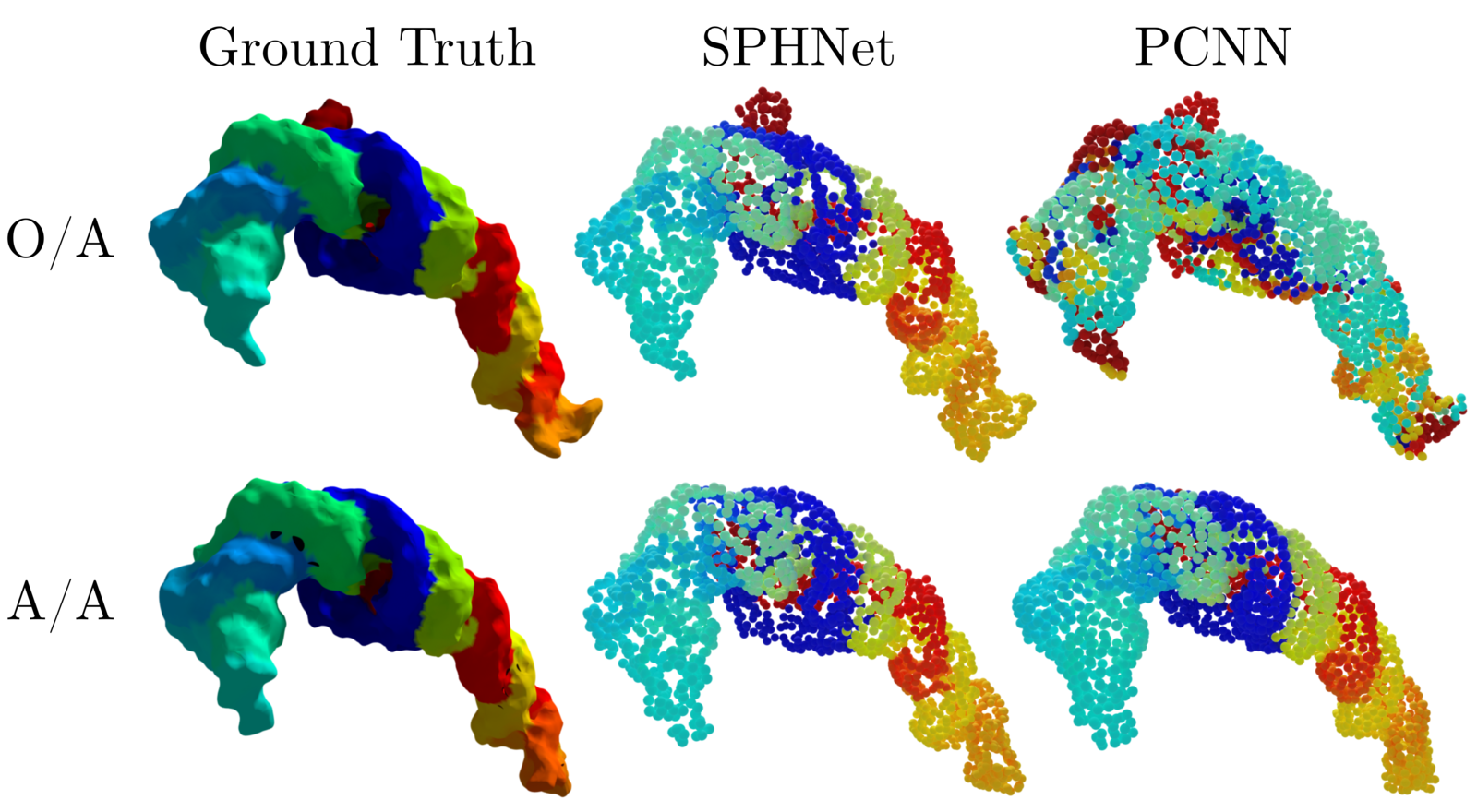}
    \caption{RNA segmentation results.}
    \label{fig:rna_segm}
\end{figure}

We also show a qualitative comparison to PCNN in Figure \ref{fig:rna_segm}. We note that when trained on the original dataset and tested on an augmented dataset, we achieve significantly better performance than PCNN. This demonstrates that unlike PCNN, the performance of our method  does not depend on the orientation of the shapes. 


\subsection{Matching}

We also apply our method on the problem of finding correspondences between non-rigid shapes. This is an especially difficult problem since in addition to \emph{global} rigid motion, the shapes can undergo non-rigid deformations, such as articulated motion of humans. 

In this setting, we trained and tested different methods on point
clouds sampled from the D-FAUST dataset \cite{dfaust}. This dataset
contains scans of 10 different subjects completing various sequences
of motions given as meshes with the same structure and indexing. We
prepared a test set consisting of 10 subject, motion sequence pairs and the complementary pairs defining our training set. Furthermore we sampled the motion sequences every 10 time-steps we selected 4068 shapes in total with a 3787/281 train/test split. 
We sampled $10$k points uniformly on the first selected mesh and then
subsampled 2048 points from them using farthest point sampling. We
then transferred  these points to all other meshes using their
barycentric coordinates in the triangles of the first mesh to have a
consistent point indexing on all point clouds. We produce labels by
partitioning the first shape in 256 Voronoi cells associated to 256
farthest point samples. We then associate a label to each cell. Our
goal then is to predict these labels and thus infer correspondences
between shape pairs. Since  this experiment is a particular instance of segmentation, we evaluate it with two metrics, first we measure the standard segmentation accuracy. In addition, to each cell $a$ we associate a cell $f(a)$ by taking the most represented cell among the predictions over $a$ and measure the average Euclidean distance between the respective centroids of $f(a)$ and the ground truth image of $a$. Table \ref{dfaust_table} shows quantitative performance of  different methods. Note that the accuracy of PointNet++ and PCNN decreases drastically when trained on the original dataset and tested on rotated (augmented) data sets. Our SPHNet performs well in all training and test settings. Moreover, SPHNet strongly outperforms existing methods with data augmentation applied both during training and testing, which more closely reflects a scenario of non pre-aligned training/test data.

\begin{figure}[t]
    \centering
     \includegraphics[width=\columnwidth]{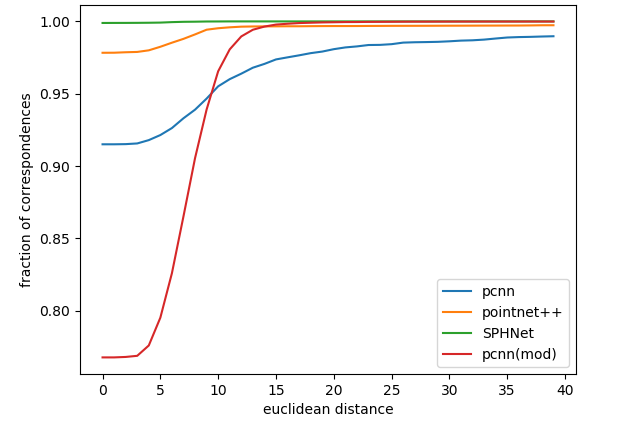}
    \caption{Fraction of correspondences on the D-FAUST dataset within a certain Euclidean error in the A/A case.}
    \label{fig:dfaust_err}
\end{figure}

\begin{figure}[t!]
    \centering
     \includegraphics[width=\columnwidth]{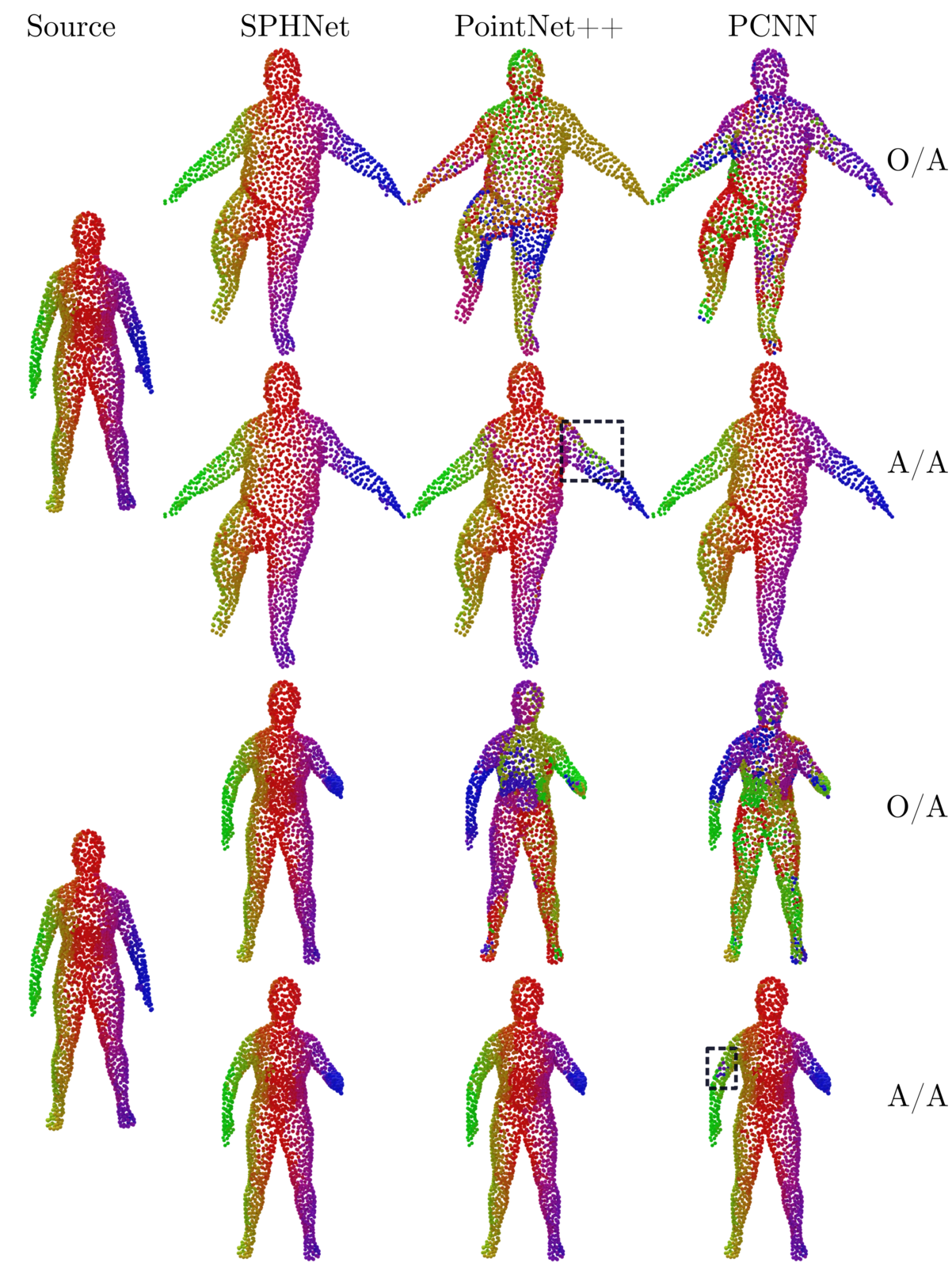}
    \caption{Qualitative comparison of correspondences on the D-FAUST dataset.}
    \label{fig:dfaust_qual2}
\end{figure}

In Figure \ref{fig:dfaust_err}, we show that the correspondences
computed by SHPNet when trained on both original and augmented data are highly accurate. For qualitative evaluation we associate a color to each Voronoi cell using the $x$ coordinate of its barycenter and transfer this color using computed correspondences.
Figure \ref{fig:dfaust_qual2}  shows a qualitative comparison of
different methods. We note that PCNN and PointNet++ correspondences present visually more artefacts including symmetry issues, while our SPHNet results in more smooth and accurate maps across all training and test settings.  

\begin{table}
\begin{center}
\begin{tabular}{|c|c|c|c|c|c|}
  \hline
  Method &  O/O & A/O &  O/A & A/A & disterr \\
  \hline
  PCNN   & \textbf{99.6} & 79.9 & 5.7 & 77.1 & $7.7e{-3}$\\
  PointNet++  & 97.1 & 85.0 & 10.4 & 84.5 & $1.8e{-3}$ \\
  PCNN (mod) & 60.4 & 55.2 & 2.5 & 54.7 & 0.01\\
  SPHNet (ours) & 98.0 & \textbf{97.2} & \textbf{91.5} & \textbf{97.1} & $\mathbf{3.5e{-5}}$\\
  PRIN & 86.7 & 3.24 &11.3  & 3.63 & 0.59 \\
  \hline
\end{tabular}
\caption{Part label prediction accuracy on our D-FAUST dataset and average Euclidean distance error of the inferred correspondences between Voronoi cell centroids in the A/A case.\label{dfaust_table}\vspace{-2mm}}
\end{center}
\end{table}

\section{Conclusion}

We presented a novel approach for ensuring rotational invariance in a point-based deep neural network, based on the previously-proposed PCNN architecture. Key to our approach is the use of spherical harmonics kernels, which are both efficient, theoretically guaranteed to be rotationally invariant and can be applied at any layer of the network, providing flexibility between local and global invariance. Unlike previous methods in this domain, our resulting network outperforms existing approaches on non pre-aligned datasets even with data augmentation. In the future, we plan to extend our method to a more general framework combining non-invariant, equivariant and fully invariant features at different levels of the network, and to devise ways for \emph{automatically} deciding the optimal layers at which invariance must be ensured. \revised{Another direction would be to apply rotationally  equivariant features across different shape segments independently. This can be especially relevant for articulated motions where different segments  undergo different but all approximately rigid motions.}
\paragraph{Acknowledgements}
Parts of this work were supported by KAUST OSR Award No. CRG-2017-3426, a gift from the NVIDIA Corporation and the ERC Starting Grant StG-2017-758800 (EXPROTEA).

{\small
\bibliographystyle{ieee}
\bibliography{egbib}
}

\section*{Appendix}
Table \ref{modelnet40_scale_table} shows the effect of the scale parameter on the accuracy of the classification task on ModelNet40. Performances decrease when $\rho$ gets too low or too high. We choose the parameter that produces the best accuracy in our experiments. 
\begin{table}
\begin{center}
\begin{tabular}{|c|c|c|c|c|c|}
  \hline
  $\rho$ &  O / O &  A / O & O / A & A / A \\
  \hline
  0.2 & 85.7 & 84.9 & 85.0 & 86.3\\
  0.15 & 86.8 & 85.6 &  86.0 & 86.7 \\
  0.1   & $\mathbf{87.7}$ & $\mathbf{87.1}$ & $\mathbf{86.6}$ &$\mathbf{87.6}$
  \\
  0.075 & 86.1 & 85.7 & 85.6 & 86.5\\
  0.05 & 85.6 & 84.5 & 83.2 & 85.4 \\
  \hline
\end{tabular}
\caption{ \revised{Impact of the scale parameter $\rho$ on classification accuracy of SPHNet on ModelNet40.} \label{modelnet40_scale_table}\vspace{-4mm}}
\end{center}
\end{table}

\end{document}